\documentclass[article]{elsarticle}

\usepackage{lineno,hyperref}
\usepackage{amsmath}
\usepackage{booktabs}
\usepackage{graphicx}
\modulolinenumbers[5]

\journal{NeuroComputing}

\begin{document}

\begin{frontmatter}

\title{Rethinking Skip Connections in Encoder-decoder Networks for Monocular Depth Estimation}
\tnotetext[mytitlenote]{Fully documented templates are available in the elsarticle package on \href{http://www.ctan.org/tex-archive/macros/latex/contrib/elsarticle}{CTAN}.}

\author[mymainaddress,mysecondaryaddress]{Zhitong Lai}

\author[mysecondaryaddress]{Haichao Sun\corref{mycorrespondingauthor}}
\ead{ciomp_shc@163.com}
\author[mysecondaryaddress]{Rui Tian}

\author[mysecondaryaddress]{Nannan Ding}

\author[mysecondaryaddress]{Zhiguo Wu}

\author[mysecondaryaddress]{Yanjie Wang}

\cortext[mycorrespondingauthor]{Corresponding author}
%\ead{ciomp_shc@163.com}

\address[mymainaddress]{Changchun Institute of Optics, Fine Mechanics and Physics, Chinese Academy of Sciences, Changchun 130033, China}
\address[mysecondaryaddress]{University of the Chinese Academy of Sciences, Beijing 100049, China}

\begin{abstract}
Skip connections are fundamental units in encoder-decoder networks, which are able to improve the feature propagtion of the neural networks. However, most methods with skip connections just connected features with the same resolution in the encoder and the decoder, which ignored the information loss in the encoder with the layers going deeper. To leverage the information loss of the features in shallower layers of the encoder, we propose a full skip connection network (FSCN) for monocular depth estimation task. In addition, to fuse features within skip connections more closely, we present an adaptive concatenation module (ACM). Further more, we conduct extensive experiments on the ourdoor and indoor datasets (i.e., the KITTI dataste and the NYU Depth V2 dataset) for FSCN and FSCN gets the state-of-the-art results.
\end{abstract}


\end{frontmatter}

%\linenumbers

\section{Introduction}

In convolutional neural networks (CNNs), convolution operation is a lossy operation. When the network layer goes deeper, the loss of information from the input increases, which causes a performance degradation. However, besides avoiding gradient exploding and gradient vanishing, skip connections are able to alleviate this degradation. 

Skip connection is first adopted in ResNet\cite{ResNet}, which is also called residual. After that, skip connections have become a popular component in many neural networks, such as WideResNet\cite{WideResNet}, DenseNet\cite{DenseNet} and ResNeXt\cite{ResNeXt}, etc. U-Net\cite{U-Net} is another type of architecture with skip connections. Different from ResNet, the skip connections in U-Net are conducted between the encoder and the decoder of an encoder-decoder architecture. Nevertheless, in this paper, we call them together as skip connections. For encoder-decoder networks, skip connections between  the encoder and the decoder are almost indispensable units due to the long distance between the input and the output. However, in most encoder-decoder networks, skip connections between encoder and decoder are just conducted between features with the same spatial resolution, which ignored the information loss of features in encoder. With the layers going deeper in encoder, the information loss of the features to be connected with decoder continues to increase. On the other hand, features of different levels in encoder contains different information, the fusion of multi-type information can be also helpful for the final prediction. In that case, we consider to connect features of all spatial levels in encoder with features of every spatial level in decoder to preserve more feature information from encoder.

To avoid large loss of feature information over a CNN, many networks have utilized dense connection mechanism. However, most of them conduct dense connections just within the encoder\cite{Chen2018,Chen2019,Xu2018} or the decoder\cite{Bilinski2018,DCPNet,Wu2021}, instead of between the encoder and the decoder. Though encoder-decoder networks contain a feature compression and a feature restoration procedure, the whole procedure of encoder-decoder can be regarded as a continuous loss of feature information. In that case, dense skip connections between encoder and decoder may make difference. 

In this paper, we call the form of dense skip connections between encoder and decoder as full skip connections. We propose a full skip connection network (FSCN) for monocular depth estimation, where the features in encoder are connected with features in decoder with a dense fashion. Monocular depth estimation based on deep neural networks (DNNs) is a pixel-wise task where the input is a single color image, and the output is a depth image at a resolution the same as the input.  Dense connection mechanism is a popular setting to alleviate information degradation and improve feature presentation ability for a CNN. For pixel-wise tasks like monocular depth estimation, propagating feature information as more as possible is vital important for more accurate predictions.

In the network FSCN, an adaptive concatenation module (ACM) is also presented, which performs better than normal concatenation in fusing features from the encoder. We conducted various experiments for monocular depth estimation task on the KITTI dataset\cite{Geiger2013} and the NYU Depth V2 dataset\cite{Silberman2012}, which get state-of-the-art results. The main contributions of our work are summarized as follows:

\begin{itemize}
\item We proposed a novel encodr-decoder network FSCN for monocular depth estimation, which employed a dense connection mechanism between the encoder and the decoder.
\item To fuse features from the encoder and features in the decoder effectively, we presented a adaptive concatenation module (ACM), which is more effective than normal concatenation.
\item We conducted extensive experiments on the KITTI datatset and the NYU Depth V2 dataset, which are outdoor dataset and indoor dataset, respectively. The results achieved the state-of-the-art on both two datasets.
\end{itemize}

The rest of this paper is organized as follows. In section \ref{sec_relw} we presented a brief review of related works. In section \ref{method}, we explained our proposed method. Extensive experiments and an ablation study are shown in section \ref{sec_exp}. In section \ref{conclusions}, we made some conlusions for our work.

\section{Related Work}
\label{sec_relw}
\subsection{Monocular Depth Estimation}
Early works for monocular depth estimation mainly focused on exploiting hand-crafted features to learn geometric or optical priors from a color image\cite{Karsch2014,Liu2010,Saxena2008}. With the development of deep learning, various DNN based models have been proposed. The methods for monocular depth estimation can be sorted as supervised fashion\cite{Eigen2014,DORN,DenseDepth,BTS}, unsupervised fashion\cite{Godard2017,Yang2020,Johnston2020,Wong2019} and semi-supervised fashion\cite{Qi2018,Yue2020,Kuznietsov2017,Ji2019}. At this stage, for the prediction accuracy, the supervised and semi-supervised fashion still have a gap with supervised fahsion. For supervised fashion, the RGB-D datasets are required to learn a mapping function to generate a corresponding depth map of a single color image. Eigen et al.\cite{Eigen2014} proposed the first DNN model for monocular depth estimation, which contains a coarse prediction stage and a refine prediction stage. Fu et al.\cite{DORN} treated monocular depth estimation as a a deep ordinal regression problem and introduced a discretization strategy. Alhashim et al.\cite{DenseDepth} introduced a model with transfer learning. Lee et al.\cite{BTS} proposed a local planar guidance module to link internel features with final output effectively, which got a great improvement. Attracted by the great success of attention machanism in capturing long-range context information, some attention-based models have been presented recently. For instance, Huynh\cite{Huynh2020} proposed a depth-attention volume (DAV) to capture more context information in the features propagation to leverage monocular depth estimation. Yang et al.\cite{TransDepth} adopted the Transformer\cite{Vaswani2017} and presented an attention gate module for monocular depth estimation. In our work, we also adopted an attention module SENet\cite{CBAM}, which was helpful for our method to get better performance.

\subsection{Skip Connections}
Skip connections are first proposed in ResNet\cite{ResNet} to solve the problem of vanishing/exploding gradients, as well as to enhance gradient propagation for deep networks, which has been one of the most fundamental elements of deep architectures. Inspired by ResNet, DenseNet\cite{DenseNet} and ResNeXt\cite{ResNeXt} were then proposed and got an improvement in parameter efficiency and feature propagation. The three architectures are usually used as a backbone network in encoder-decoder architectures. Nevertheless, as the success shown in U-Net\cite{U-Net}, skip connections between encoder and decoder can be also helpful for parameter efficiency and feature propagation. For instance, Collin et al.\cite{Collin2020} proposed an autoencoder network with skip connections to leverage anomaly detection. Bulat et al.\cite{Bulat2020} proposed a hybrid network combining the HourGlass and U-Net architectures, in which the soft-gated skip connections were presented and made great difference for human pose estimation. Wang et al.\cite{Wang2019} proposed a fully convolutional neural network with long and short skip connections for monocular depth estimation while performed well. 

To further utilize the advantages of skip connections, besides some backbones like DenseNet, many works adopted a dense fashion for skip connections. For example, Shang et al.\cite{Shang2020} proposed a novel CNN for SAR image classification, in which the dense connections were used to reuse feature maps and strengthen information transmission. Dai et al.\cite{Dai2021} proposed a dense scale network for crowd counting, which is an encoder-decoder architecture and the decoder contained a dense skip connection mechanism. Bao et al.\cite{Bao2020} proposed a multi-scale residual dense network that employed dense connection many times for image denoising.

For encoder-decoder architectures, skip connections act as a highway to pass details of the previous feature maps in the encoder to the decoder. However, most skip connections just pass feature maps in the encoder to features with the same resolution in the decoder, which ignored the details of features with other resolutions in the encoder. In that case, we construct highways for all spatial levels of features in the encoder with every spatial level of features in the decoder, which forms a dense fashion.

\section{Method}
\label{method}
In this section, we first introduce our proposed network FSCN, then describe the details of adaptive concatenation module (ACM). Finally, we introduce the loss function for the training procedure.

\subsection{Network Architecture}

\begin{figure}[h]
	\centering
	\includegraphics[width=12 cm]{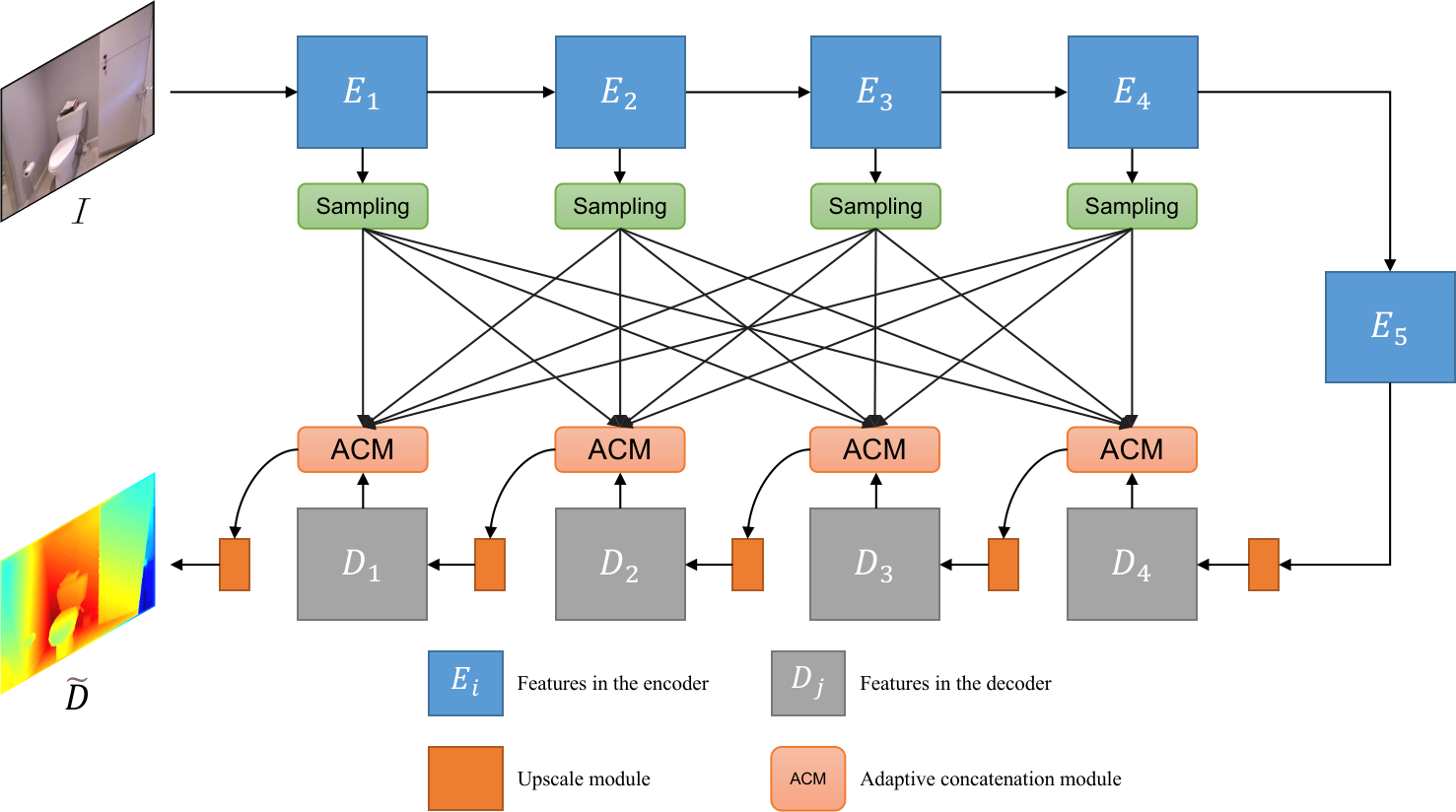}
	\caption{The overall architecture of FSCN, in which $E_{i}$ $(i \in 0 \sim 5)$ and $D_{j}$ $(j \in 0 \sim 4)$ indicate features with different spatial resolutions in the encoder and the decoder, respectively. Note that when $i=j$, the shapes of $E_{i}$ and $D_{j}$ are the same.}
	\label{architecture}
\end{figure}

The overall architecture of our proposed network is shown in Figure \ref{architecture}. The FSCN is an encoder-decoder architecture. Except the densest feature extracted by the encoder (i.e., $E_{5}$ in the figure), the features in all spatial levels of the encoder are preserved to be concatenated with features in the decoder. Before the concatenation, the features in the encoder are first scaled to specific spatial resolutions that are the same as the ones of features in the decoder by sampling operations. Then the scaled features are concatenated with features in the decoder by adaptive concatenation modules (i.e., ACM in the figure), which will be described in Section \ref{sec:ACM}. After the concatenation, the features are sent to a SENet\cite{CBAM} module and then upscaled to next spatial level by an upscale operation with a ratio of 2. Finally, the network outputs a depth map with a resolution the same as the input image.

The upscale module contains a sequence with two elements, which are a upsampling operation and a convolution operation, the former is to upscale the spatial resolution of a feature with a ratio of 2, while the latter is to alter the channel number into next level.

\subsection{Adaptive Concatenation Module}\label{sec:ACM}

\begin{figure}[h!]
	\centering
	\includegraphics[width=12 cm]{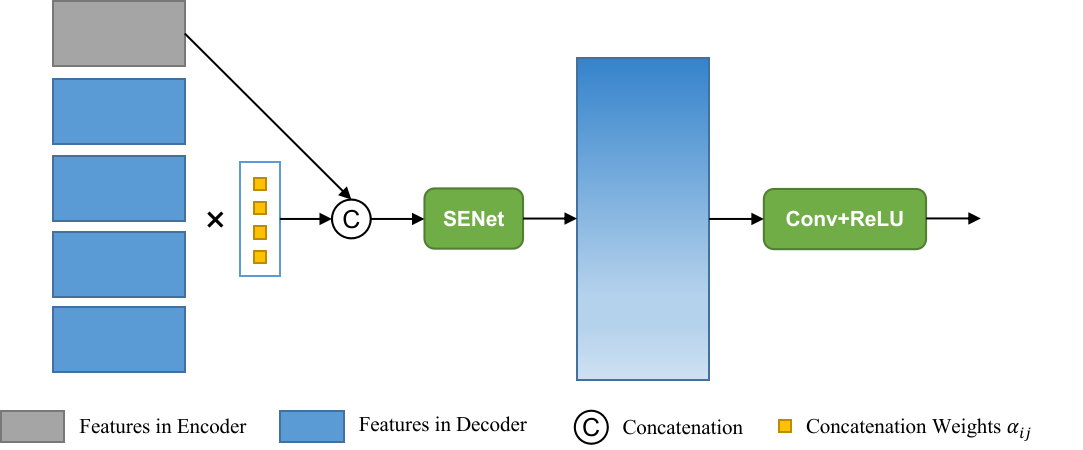}
	\caption{The overall architecture of adaptive concatenation module (ACM)}
	\label{concat}
\end{figure} 

As shown in Figure \ref{concat}, we design an adaptive concatenation module to fuse features from the encoder with features in the decoder adaptively. In Figure \ref{concat}, $D_{j}$ $(j \in [1, 4])$ denotes the features in specific spatial levels of the decoder, $E_{ij}$ are features altered from features $E_{i}$ by sampling operations, with whose spatial resolution the same as $D_{j}$. $\alpha_{ij}$ are a series of learnable parameters with whose initial values are a random number on the interval [0, 1), which are used to decide the importance of each feature block from the encoder when concatenated with features in the decoder. We call $\alpha_{ij}$ concatenation weights in this paper. In Figure \ref{concat}, $\copyright$ indicates concatenation operation, the procedure can be presented as equation \ref{eq: concat}.

\begin{equation}\label{eq: concat}
	D_{j}^{'}=Concat(\alpha_{1j}E_{1j}, \alpha_{2j}E_{2j}, \alpha_{3j}E_{3j}, \alpha_{4j}E_{4j}, D_{j})
\end{equation}

Though a random ratio is used before the concatenation, the weights are not set in channel-wise. As the channel number of the feature obtained from the concatenation increases intensely, we consider to employ an SENet\cite{CBAM} for the concatenated feature block. SENet is a channel attention module that is able to decide the weight of each channel of a feature, which can advance the presentation ability of a feature.

After an SENet module, a convolution operation is used to fuse the feature after the concatenation, then we get the output $F_{j}$. Note that the shape of $F_{j}$ is the same as feature $D_{j}$. This procedure can be presented as equation \ref{eq: adapt}.

\begin{equation}\label{eq: adapt}
	F_{j} = ReLU(Conv(SENet(D_{j}^{'})))
\end{equation}

Since the ratios $\alpha_{ij}$ and channel attention module SENet are used in equation \ref{eq: concat} and equation \ref{eq: adapt}, respectively. We call the procedures consisted by equation \ref{eq: concat} and equation \ref{eq: adapt} adaptive concatenation module. In Section \ref{sec:exp}, several experiments will be conducted for proving the effectiveness of our proposed adaptive concatenation module.

\subsection{Loss Function}

We adopt the improved scale-invariant loss function introduced in \cite{BTS} for the training phase. The scale-invariant loss is proposed by Eigen et al.\cite{Eigen2014}, which is:

\begin{equation}\label{scal_inv}
L(y, y^{*})=\frac{1}{N} \sideset{}{_i}\sum d^2_{i} - \frac{\lambda}{N^2} (\sideset{}{_i}\sum d_{i})^2
\end{equation}
in which $d_{i}=logy_{i}-log\hat{y_{i}}$, $y_{i}$ and $\hat{y_{i}}$ denote the ground truth and predicted depth map at pixel $i$, respectively. $N$ denotes the total pixels of a depth map. The improved scale-invariant loss is:

\begin{equation}\label{imp_scal_inv}
L=\alpha\sqrt{L(y, y^{*})}
\end{equation}
The hyper-parameter $\alpha$ is set to 10 and $\lambda$ is set to 0.85, which is the same as the ones in [21]. 

\section{Experiments}\label{sec:exp}
\label{sec_exp}
Multiple experiments for evaluation are conducted on two baseline datasets, i.e., the KITTI dataset and the NYU Depth V2 dataset. Both quantitative and qualitative results are provided. Moreover, we set some comparisons with other representative monocular depth estimation method, i.e., \cite{Eigen2014,Liu2015,DenseDepth,DORN,Yin2019,BTS,Godard2019,TransDepth,Liu2021,Ye2021,Hu2019,Chen_arx2019,Xu2021}.

\subsection{Datasets}
The KITTI dataset\cite{Geiger2013} is a large-scale outdoor dataset captured by multiple sensors mounted on a driving car, which is created for automatic driving researches. The dataset contains a number of color images and corresponding depth maps with a resolution of 375 $\times$ 1242 pixels. For the experiments, we adopted a data spilt strategy proposed by Eigen et al.\cite{Eigen2014}, in which the training set contains 23,488 images from 32 scenes and the test set contains 697 images from remaining 29 scenes. The images are cropped to 352 $\times$ 704 in a random manner in the experiments.

The NYU Depth V2 dataset\cite{Silberman2012} is an indoor dataset containing 120K RGB images and paired depth maps from 464 indoor scenes. The resolusion of the color images and depth maps is 480 × 640 pixels. For the experiments, we crop the images into 416 $\times$ 544 pixels randomly. The training set and test set are also splited with the strategy proposed by Eigen et al.\cite{Eigen2014}, in which the training set contains 36253 pairs from 249 scenes and the test set contains 654 pairs from 251 scenes.

\subsection{Evaluation Metrics}
To evaluate the performance of our method, we adopted standard evaluation metrics used in previous works\cite{Eigen2014,BTS,TransDepth}:
\begin{itemize}
	\item	Mean relative error ($abs\ rel$): $\frac{1}{N} \sideset{}{_{i=1}^N}\sum \frac{\Vert \hat{d_{i}}-d_{i} \Vert}{d_{i}}$;
	\item	Squared relative error ($sq\ rel$): $\frac{1}{N} \sideset{}{_{i=1}^N}\sum \frac{\Vert \hat{d_{i}}-d_{i} \Vert^2}{d_{i}}$;
	\item	Root mean squared error ($rms$): $\sqrt{\frac{1}{N} \sideset{}{_{i=1}^N}\sum (\hat{d_{i}}-d_{i})^2}$;
	\item	Mean log10 error ($log10$): $\frac{1}{N} \sideset{}{_{i=1}^N}\sum \Vert log\hat{d_{i}}-logd_{i} \Vert$;
	\item	Root mean squared log10 error ($log\ rms$): $\sqrt{\frac{1}{N} \sideset{}{_{i=1}^N}\sum \Vert log \hat{d_{i}}-log d_{i} \Vert^2}$;
	\item	Accuracy with threshold $\tau$, i.e., the percantage (\%) of $\hat{d_{i}}$ subjecting to $\delta=max(\frac{d_{i}}{\hat{d_{i}}}, \frac{\hat{d_{i}}}{d_{i}})<1.25^{\tau}$, here, $\tau \in (1, 2, 3)$.
\end{itemize}
where $N$ indicates the total number of valid pixels in the ground truth. $y_{i}$ and $\hat{y_{i}}$ denote ground truth and predicted depth value at pixel $i$, respectively.

\subsection{Implementation Details}
We implemented all the experiments with the open source deep learning framework PyTorch. Two NVIDIA 3090 GPUs are used for all trainings. When training, we employed the AdamW optimizer\cite{Glorot2010} with $\beta_{1}=0.9$, $\beta_{2}=0.999$ and $\epsilon=10^{-6}$. The number of epochs was set to 50. The batch size was set to 8. We initialize the weights with Xavier initialization\cite{Deng2009}. The initial leraning rate was set to $10^{-4}$ and decayed with the strategy proposed in \cite{BTS}.

We chose the backbone DenseNet161\cite{DenseNet} that was pretrained on ImageNet\cite{Simonyan2014} for the encoder part of our network. Moreover, we conducted expriments for ablation study of three concatenation methods. We also explored the influences of discarding specific skip connections from specific features in the encoder.

We employed data augmentations to improve training performance and avoid overfitting. The augmentations include random horizontal flipping, random contrast, random color adjustment with a chance of 50\%. Random rotation was also used, with the angles in range of [-1, 1] for the KITTI dataset, and [-2.5, 2.5] for the NYU Depth V2 dataset.

\subsection{Results on the KITTI Dataset}
Table \ref{exp:KITTI} shows the quantitative results of our proposed method on the KITTI Eigen split. Note that \cite{Eigen2014}\cite{BTS}\cite{Ye2021}\cite{Godard2019}\cite{TransDepth} employed the same split strategy as our method.

\begin{table}[h!]
	\caption{Experimental results on the KITTI Eigen split. The values in bold type are the best results of every metric among these works, while the values underlined are the second best results. We set the depth range to 0-80m. Metrics marked by $\downarrow$: lower is better; metrics marked by $\uparrow$: higher is better.}
	\label{exp:KITTI}
	\centering
	\resizebox{\textwidth}{!}{
		\begin{tabular}{rccccccc}
			\toprule
			\textbf{Method}	 & \textbf{$abs\ rel$}$\downarrow$	& \textbf{$sq\ rel$}$\downarrow$ & \textbf{$rms$}$\downarrow$ & \textbf{$log\ rms$}$\downarrow$ & \textbf{$\delta < 1.25$}$\uparrow$ & \textbf{$\delta < 1.25^{2}$}$\uparrow$ & \textbf{$\delta < 1.25^{3}$}$\uparrow$\\
			\midrule
			Eiegn et al.\cite{Eigen2014}	    & 0.190 			& 1.515 			& 7.156 			& 0.270 			& 0.692 			& 0.899 			& 0.967\\
			Liu et al.\cite{Liu2015}		    & 0.217 			& - 				& 7.046 			& - 				& 0.656 			& 0.881 			& 0.958\\
			DenseDepth\cite{DenseDepth}    & 0.093 			& 0.589 			& 4.170 			& - 				& 0.886 			& 0.965 			& 0.986\\
			DORN\cite{DORN}		        & 0.072 			& 0.307 			& \bfseries{2.727}  & 0.120 			& 0.932 			& 0.984 			& 0.994\\
			Yin et al.\cite{Yin2019}		    & 0.072 			& - 				& 3.258 			& 0.117 			& 0.938 			& 0.990 			& \underline{0.998}\\
			BTS\cite{BTS}		        & \bfseries{0.060}  & \underline{0.249} & 2.798 		    & \bfseries{0.096}  & \underline{0.955}  & 0.993             & \underline{0.998}\\
			Godard et al.\cite{Godard2019}	    & 0.106 			& 0.806 			& 4.530 			& 0.193 			& 0.876 			& 0.958 			& 0.980\\
			TransDepth\cite{TransDepth}        & 0.064 			& 0.252			    & 2.755 & 0.098             & \bfseries{0.956}  & \bfseries{0.994}  & \bfseries{0.999}\\
			Liu et al.\cite{Liu2021}		    & 0.111 			& - 				& 3.514 			& - 				& 0.878 			& 0.977 			& 0.994\\
			DPNet\cite{Ye2021}		        & 0.112 			& - 				& 4.978 			& 0.210 			& 0.842 			& 0.947 			& 0.973\\
			WaveletMnodepth\cite{ramamonjisoa2021single}  & 0.097 			& 0.718 				& 4.387 			& 0.184 			& 0.891 		&0.962 			& 0.982\\
			\midrule
			FSCN 	                & \underline{0.062} & \bfseries{0.248} 	& \underline{2.739} 			& \underline{0.097}	& \underline{0.955} & \underline{0.993}	& \bfseries{0.999}\\
			\bottomrule
		\end{tabular}
	}
\end{table}

From Table \ref{exp:KITTI} we can see that our method gets competitive results with current leading algorithms. On metric $rms$, our method works worser than DORN, however, it performs much better than other algorithms. Moreover, our method works much better than DORN except metric $rms$.

\begin{figure}[h!]
	\centering
	\includegraphics[width=12 cm]{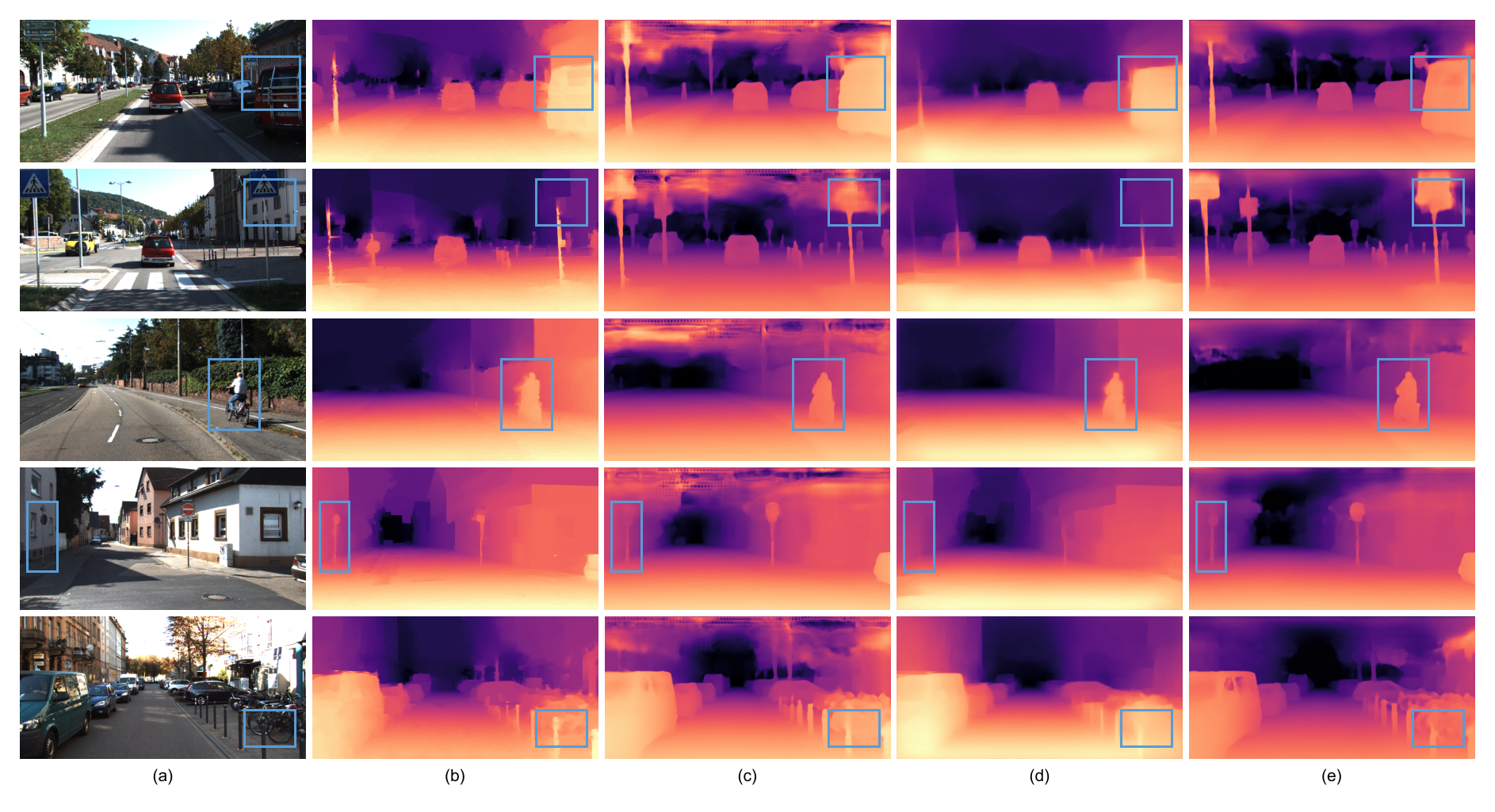}
	\caption{Qualitative examples on the KITTI Eigen test split. (a) RGB image; (b) ground truth; (c) BTS\cite{BTS}; (d) DenseDepth\cite{DenseDepth}; (e) our proposed FSCN. The ground truth depth maps are filled based on sparse point clouds utilizing tools provided by the NYU Depth V2 dataset. For better visualization, the values of all the depth maps are logarithmic. Note that the encoders of BTS\cite{BTS} and FSCN are both DenseNet161.}
	\label{fig:exp_kitti}
\end{figure} 

Figure \ref{fig:exp_kitti} shows the qualitative results of FSCN on the KITTI Eigen validation set, while comparing with two leading algorithms BTS\cite{BTS} and DenseDepth\cite{DenseDepth}. From this figure, we can observe that FSCN method shows more details in the contents like cars structure, traffic signs, sketch of human and so on, comparing with the other two counterpart methods, which may convey the evidence that our method with full skip connection mechanism is able to preserve and propagate more feature information along the deep network.

\subsection{Results on the NYU Depth V2 Dataset}

\begin{table}[h!]
	\caption{Experimental results on the NYU Depth V2 Eigen split. The values in bold type are the best results of every metric among these works, while the values underlined are the second best results. We set the depth range to 0-10m. Metrics marked by $\downarrow$: lower is better; metrics marked by $\uparrow$: higher is better.}
	\label{exp:NYU}
	\centering
	\resizebox{\textwidth}{!}{
		\begin{tabular}{rccccccc}
			\toprule
			\textbf{Method}	& \textbf{$abs\ rel$}$\downarrow$	& \textbf{$log10$}$\downarrow$ & \textbf{$rms$}$\downarrow$ & \textbf{$\delta < 1.25$}$\uparrow$ & \textbf{$\delta < 1.25^{2}$}$\uparrow$ & \textbf{$\delta < 1.25^{3}$}$\uparrow$\\
			\midrule
			Eiegn et al.\cite{Eigen2014}	& 0.215 			& -		 			& 0.907 			& 0.611 			& 0.887 			& 0.971\\
			Liu et al.\cite{Liu2015}		& 0.213 			& 0.087 			& 0.759 			& 0.650 			& 0.906 			& 0.976\\
			Fu et al.\cite{DORN}		    & 0.115 			& 0.051 			& 0.509				& 0.828 			& 0.965 			& 0.992\\
			Hu et al.\cite{Hu2019}		    & 0.123 			& 0.053 			& 0.544			 	& 0.855 			& 0.972 			& 0.993\\
			Yin et al.\cite{Yin2019}		& \underline{0.108} & 0.048 & \underline{0.416} & 0.875             & 0.976 			& \underline{0.994}\\
			Chen et al.\cite{Chen_arx2019}  & 0.111             & \underline{0.048}	& 0.514 			& \underline{0.878} & \underline{0.977}	& \underline{0.994}\\
			Liu et al.\cite{Liu2021}		& 0.113 			& 0.049 & 0.525 			& 0.872 			& 0.974 			& 0.993\\
			Ye et al.\cite{Ye2021}	        & -		 			& 0.063 			& 0.474 			& 0.784 			& 0.948 			& 0.986\\
			Xu et al.\cite{Xu2021}		    & \bfseries{0.101}  & 0.054 			& 0.456 			& 0.823 			& 0.962 			& \underline{0.994}\\
			WaveletMnodepth\cite{ramamonjisoa2021single}  & 0.126 			& 0.054 				& 0.552 			& 0.845 			& 0.968 		&0.992 \\
			\midrule
			
			FSCN				& 0.111 			& \bfseries{0.047}  & \bfseries{0.395}  & \bfseries{0.884}  & \bfseries{0.981}	& \bfseries{0.995}\\
			\bottomrule
		\end{tabular}
	}
\end{table}

Table \ref{exp:NYU} shows the quantitative results of FSCN network on the NYU Depth V2 dataset. Comparing with other methods in this table, FSCN network performs best except in the metric $abs\ rel$. Especially, our method performs much better than other methods in metric $rms$ and $\delta < 1.25^{2}$.

\begin{figure}[h!]
	\centering
	\includegraphics[width=12 cm]{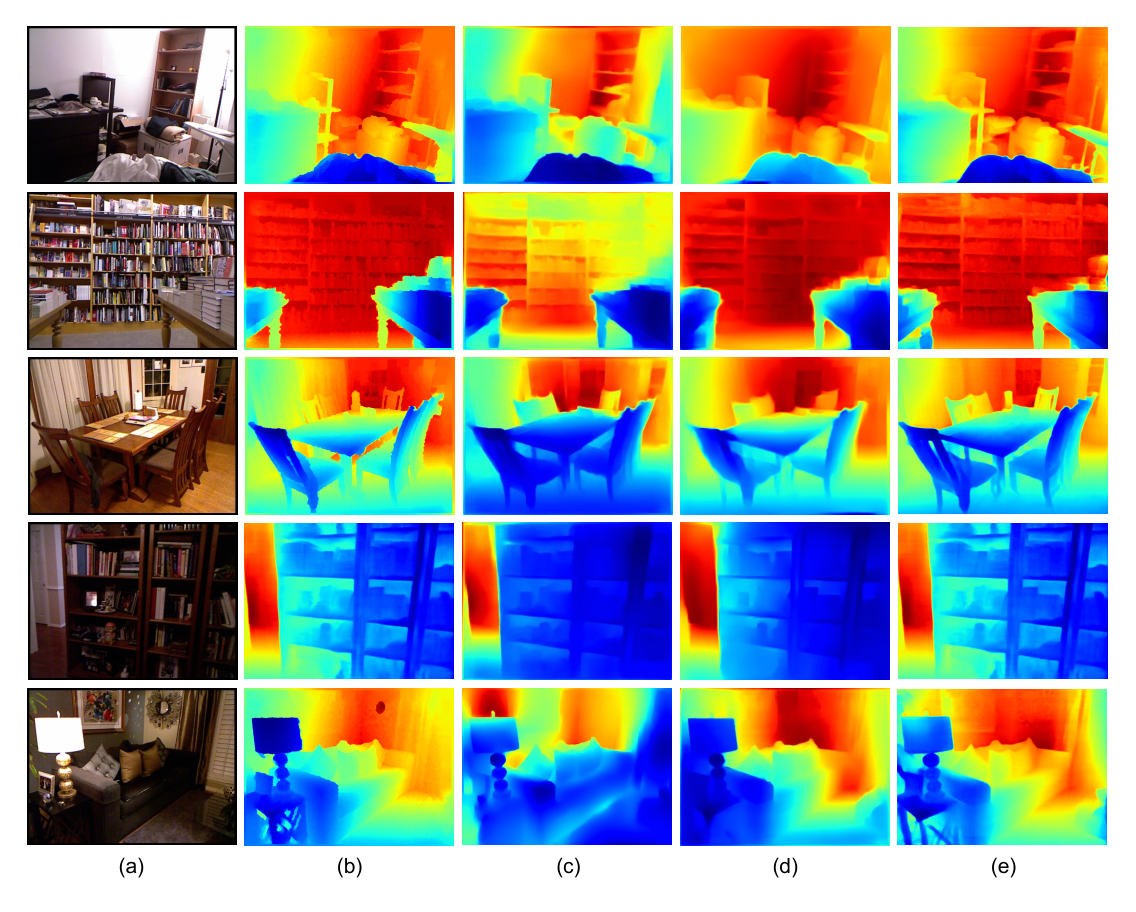}
	\caption{Qualitative examples on the NYU Depth V2 Eigen test split. (a) RGB image; (b) ground truth; (c) Hu et al.\cite{Hu2019}; (d) Chen et al.\cite{Chen_arx2019}; (e) Ours. From top to bottom, We select five RGB images from five scenes, i.e., bedroom, bookstore, dining room, home office and kitchen, respectively. Note that the encoder of Ours is DenseNet161.}
	\label{fig:exp_nyu}
\end{figure} 

Figure \ref{fig:exp_nyu} shows the qualitative results of FSCN working on the NYU Depth V2 dataset, from which we can observe that FSCN network performs excellent in predicting the details like shelves and chair legs. The comparisons in Figure \ref{fig:exp_nyu} prove the effectiveness of our proposed FSCN network.

\subsection{Ablation Study}
\subsubsection{Effect of Full Skip Connections}

We implemented an ablation study to explore the effect of full skip connections conducted in our method. We set a comparison among three setups, which are no skip connection (indicated as "no-skip"), skip connections conducted within the same spatial level between the encoder and the decoder (indicated as "same-skip") and full skip connections introduced in this paper (indicated as "full-skip"). We implement experiments both on the  KITTI dataset and the NYU Depth V2 dataset. For equality we preserve the adaptive concatenation module in "same-skip" counterpart.

\begin{table}[h]
	\caption{Experimental results on the KITTI Eigen split for different skip-connection mechanisms. We set the depth range to 0-80m. Metric \#params means the total number of parameters of specific experimental setups. Metrics marked by $\downarrow$: lower is better; metrics marked by $\uparrow$: higher is better.}
	\label{abs1:kitti}
	\centering
	\resizebox{\textwidth}{!}{
		\begin{tabular}{rcccccccc}
			\toprule
			\textbf{Method}	 & \#params & \textbf{$abs\ rel$}$\downarrow$	& \textbf{$sq\ rel$}$\downarrow$ & \textbf{$rms$}$\downarrow$ & \textbf{$log\ rms$}$\downarrow$ & \textbf{$\delta < 1.25$}$\uparrow$ & \textbf{$\delta < 1.25^{2}$}$\uparrow$ & \textbf{$\delta < 1.25^{3}$}$\uparrow$\\
			\midrule
			no-skip				& 35.04M 	& 0.063 & 0.252	& 2.793 			& 0.099	& 0.953 & 0.993	& 0.998\\
			same-skip			& 38.77M	& 0.062 & 0.246 	& 2.787 			& 0.098	& 0.954 & 0.993	& 0.999\\

			FSCN (full-skip)	  & 42.62M         & 0.062 & 0.248	& 2.739 			& 0.097	& 0.955 & 0.993	& 0.999\\
			\bottomrule
		\end{tabular}
	}
\end{table}

\begin{table}[h]
	\caption{Experimental results on the NYU Depth V2 Eigen split for different skip-connection mechanisms. The values in bold type are the best results of every metric among these works, while the values underlined are the second best results. We set the depth range to 0-10m. Metric \#params means the total number of parameters of specific experimental setups. Metrics marked by $\downarrow$: lower is better; metrics marked by $\uparrow$: higher is better.}
	\label{abs1:nyu}
	\centering
	\resizebox{\textwidth}{!}{
		\begin{tabular}{rcccccccc}
			\toprule
			\textbf{Method}	& \#params & \textbf{$abs\ rel$}$\downarrow$	& \textbf{$log10$}$\downarrow$ & \textbf{$rms$}$\downarrow$ & \textbf{$\delta < 1.25$}$\uparrow$ & \textbf{$\delta < 1.25^{2}$}$\uparrow$ & \textbf{$\delta < 1.25^{3}$}$\uparrow$\\
			\midrule
			no-skip				    & 35.04M 			& 0.113 			& 0.049  & 0.404  & 0.876  & 0.981	& 0.996\\
			same-skip			 & 38.77M		& 0.112 			& 0.048  & 0.397  & 0.878  & 0.980	& 0.996\\
			Ours (full-skip)	& 42.62M		& 0.111 			& 0.047  & 0.395  & 0.884  & 0.981	& 0.995\\
			\bottomrule
		\end{tabular}
	}
\end{table}

The experimental results on the KITTI dataset and the NYU Depth V2 dataset are shown in Table \ref{exp:KITTI} and Table \ref{exp:NYU}, respectively. From the two tables we can observe that our method with full skip connections performs better than the counterparts "same-skip" and "no-skip", which proves the effectiveness of full skip connection mechanism utilized in FSCN network. Moreover, the setup "same-skip" works better than the setup "no-skip", which shows the advantage of skip connection mechanism in CNNs. Interestingly, the setup "no-skip" performs better than many methods listed in Table \ref{exp:KITTI} and Table \ref{exp:NYU}, which offers us a direction for future research work.

\subsubsection{Effect of Adaptive Concatenation Module}

Adaptive concatenation module is an important part in FSCN network. The reason why it is called "adaptive" is because two items within it, i.e., the concatenation weights (CW) and channel attention module SENet (SE). In this section, we set several experiments to evaluate the effect of ACM, which are counterparts discarding concatenation weights (CW), counterparts discarding SENet (SE) and counterparts discarding both items. All the experiments are implemented on both the KITTI dataset and the NYU Depth V2 dataset.

\begin{table}[h!]
	\caption{Experimental results on the KITTI Eigen split for different setups of ACM. We set the depth range to 0-80m.  Metric \#params means the total number of parameters of specific experimental setups. Metrics marked by $\downarrow$: lower is better; metrics marked by $\uparrow$: higher is better.}
	\label{abs2:kitti}
	\centering
	\resizebox{\textwidth}{!}{
		\begin{tabular}{rcccccccc}
			\toprule
			\textbf{Method}	 & \#params & \textbf{$abs\ rel$}$\downarrow$	& \textbf{$sq\ rel$}$\downarrow$ & \textbf{$rms$}$\downarrow$ & \textbf{$log\ rms$}$\downarrow$ & \textbf{$\delta < 1.25$}$\uparrow$ & \textbf{$\delta < 1.25^{2}$}$\uparrow$ & \textbf{$\delta < 1.25^{3}$}$\uparrow$\\
			\midrule
			w/o CW			& 42.62M 	& 0.062 & 0.249	& 2.749 			& 0.097	& 0.954 & 0.993	& 0.999\\
			w/o SE			& 42.15M 	& 0.062 & 0.251	& 2.818 			& 0.099	& 0.953 & 0.993	& 0.998\\
			w/o CW\&SE	  & 42.15M         & 0.064 & 0.257	& 2.837 			& 0.100	& 0.952 & 0.993	& 0.998\\
			FSCN	  & 42.62M         & 0.062 & 0.248	& 2.739 			& 0.097	& 0.955 & 0.993	& 0.999\\
			\bottomrule
		\end{tabular}
	}
\end{table}

\begin{table}[h!]
	\caption{Experimental results on the NYU Depth V2 Eigen split for different setups of ACM. The values in bold type are the best results of every metric among these works, while the values underlined are the second best results. We set the depth range to 0-10m. Metric \#params means the total number of parameters of specific experimental setups. Metrics marked by $\downarrow$: lower is better; metrics marked by $\uparrow$: higher is better.}
	\label{abs2:nyu}
	\centering
	\resizebox{\textwidth}{!}{
		\begin{tabular}{rcccccccc}
			\toprule
			\textbf{Method}	& \#params & \textbf{$abs\ rel$}$\downarrow$	& \textbf{$log10$}$\downarrow$ & \textbf{$rms$}$\downarrow$ & \textbf{$\delta < 1.25$}$\uparrow$ & \textbf{$\delta < 1.25^{2}$}$\uparrow$ & \textbf{$\delta < 1.25^{3}$}$\uparrow$\\
			\midrule
			w/o CW			    & 42.62M 			& 0.112 			& 0.048  & 0.397  & 0.881  & 0.981	& 0.995\\
			w/o SE			& 42.15M 	& 0.114 & 0.048	& 0.403 			& 0.874	& 0.977 & 0.994	\\
			w/o CW\&SE	& 42.15M		& 0.114			& 0.049 & 0.408  & 0.875  & 0.978	& 0.995\\
			FSCN	& 42.62M		& 0.111 			& 0.047  & 0.395  & 0.884  & 0.981	& 0.995\\
			\bottomrule
		\end{tabular}
	}
\end{table}

The experimental results in Table \ref{abs2:kitti} and Table \ref{abs2:nyu} show that the performance of the network drops dramatically when discarding the concatenation weights or SENet, which proves the effectiveness of our proposed adaptive concatenation module. The comparison between "w/o CW" and "w/o SE" shows that SENet plays a more important role in FSCN than concatenation weights. Moreover, the idea of ACM can be transferred into other networks.

\section{Conclusions}
\label{conclusions}

In this work, we proposed a so called full skip connection based encoder-decoder network for monocular depth estimation. Comparing with traditional skip connections in normal encoder-decoder networks, the full skip connections presented in our work leveraged the information loss of feature propagation in deep networks. Moreover, we presented an adaptive concatenation module to fuse the features to be connected. Our proposed method achieved state-of-the-art results on the KITTI dataset and the NYU Depth V2 dataset, which demonstrated the effectiveness of our method for monocular depth estimation task. The ablation study proved the effectiveness of the presented full skip connection mechanism and adaptive concatenation module, which offered a new idea for us to utilize skip connections within CNNs.

\end{document}